# BengaliFig: A Low-Resource Challenge for Figurative and Culturally Grounded Reasoning in Bengali

**Abdullah Al Sefat**
Independent Researcher
abd.al.sefat@gmail.com

## Abstract

Large language models excel on broad multilingual benchmarks but remains to be evaluated extensively in figurative and culturally grounded reasoning, especially in low-resource context. We present **BengaliFig**, a compact yet richly annotated challenge set that targets this gap in Bengali, a widely spoken low-resourced language. The dataset contains 435 unique riddles drawn from Bengali oral and literary traditions. Each item is annotated along five orthogonal dimensions capturing reasoning type, trap type, cultural depth, answer category, and difficulty, and is automatically converted to multiple–choice format through a constraint–aware, AI–assisted pipeline. We evaluate eight frontier LLMs from major providers under zero–shot and few–shot chain–of–thought prompting revealing consistent weaknesses in metaphorical and culturally specific reasoning. BengaliFig thus contributes both a diagnostic probe for evaluating LLM robustness in low-resource cultural contexts and a step toward inclusive and heritage-aware NLP evaluation. Data and evaluation code is available at https://github.com/chaoSefat/Bengali-Fig

## 1 Introduction

Over the years we have seen several largescale Question-Answer(QA) datasets such as SQuAD (Rajpurkar et al., 2016), TriviaQA (Joshi et al., 2017) and Natural Questions (Rajpurkar et al., 2016). Datasets such as DROP (Dua et al., 2019), ARC (Clark et al., 2018) and MMLU (Hendrycks et al., 2021) push models towards deeper knowledge and structured reasoning skills rather than simple literal QA. Large Language Models (LLMs) have achieved impressive results on such large-scale benchmarks. However, figurative, metaphorical and culturally grounded reasoning are blindspots of these large scale datasets. While some work has been done in metaphor detection (Leong et al., 2020), (Maudslay et al., 2020), (Lu and Wang, 2017), (Wang et al., 2019),the focus is on high-resourced languages such as Chinese and English, leaving many widely spoken but under-resourced languages unexplored.

Small, focused probe tasks have proven useful for diagnosing specific reasoning capabilities (e.g., the Winograd Schema Challenge (Levesque et al., 2012), HANS (McCoy et al., 2019), StressTest (Naik et al., 2018)). Such resources demonstrate that fewer deliberately curated examples can reveal failure modes that large corpora and benchmarks do not reveal. This is particularly important for low-resource languages, where cultural and oral traditions encode figurative reasoning that is rarely captured by existing datasets.

Riddles are an oral and literary form rich in metaphor, misdirection, and local knowledge form a natural diagnostic arena but are absent from current evaluation suites. Bengali figurative riddles often encode perceptual and symbolic cues, referencing color, form, sound, and motion and thus offering a textual lens into reasoning that naturally spans multiple modalities. Bengali is the $7^{th}$ most spoken language in the world[1], yet no evaluation specifically probes figurative or culturally grounded reasoning in Bengali.

To address this gap we present **BengaliFig**, a challenge set crafted to stress-test figurative reasoning and cultural grounding in Bengali. Our contributions are threefold:

1. **Challenge set creation:** We curate and release a corpus of 435 unique Bengali riddles, each manually deduplicated, normalized and structured as Multiple Choice Question (MCQ) format.

2. **Multi-axis Annotation:** We annotate our curated QA dataset over five orthogonal di-

---
[1] https://www.statista.com/statistics/266808/the-most-spoken-languages-worldwide/

mensions capturing cognitive and cultural attributes.

3. **Comprehensive evaluation:** We probe eight frontier LLMs under zero-shot and few-shot chain-of-thought prompting. We then analyze their performance breakdown over the annotated dimensions and prompting techniques.

Our results demonstrate that majority of the frontier LLMs struggle significantly with Bengali riddles. BengaliFig thus fills a critical gap by providing a culturally grounded, low-resource testbed for probing LLM robustness and for guiding more inclusive NLP research.

## 2 Related Works

Multilingual benchmarks such as FLORES-200 (Team et al., 2022), XTREME (Hu et al., 2020), and IndicGLUE (Kakwani et al., 2020) include Bengali but focus primarily on translation, classification, or factual QA. Dedicated Bengali resources include BanglaNLG for natural language generation (Bhattacharjee et al., 2023), BanglaRQA for reading comprehension (Ekram et al., 2022), Vashantor for dialect translation (Faria et al., 2023), BenNumEval for numerical reasoning (Ahmed et al., 2025), and BEnQA for middle- and high-school QA (Shafayat et al., 2024). These tasks remain largely literal and do not assess figurative, metaphorical, or culturally embedded reasoning.

Research on figurative language has focused primarily on high-resource languages such as English and Chinese. Prior work includes metaphor detection (Leong et al., 2020; Maudslay et al., 2020; Lu and Wang, 2017; Wang et al., 2019) and broader figurative understanding (Jang et al., 2023; Lai and Nissim, 2024). Riddle-focused resources such as BiRdQA (Zhang and Wan, 2021), CC-Riddles (Xu et al., 2023), and Visual Riddles (Bitton-Guetta et al., 2024) probe models' ability to integrate metaphor, ambiguity, and cultural knowledge. However, these datasets remain concentrated in high-resource languages and do not extend to Bengali.

Despite Bengali being one of the world's most widely spoken languages, no benchmark targets metaphorical, figurative, or culturally grounded reasoning. Such abilities are deeply rooted in cultural context, making Bengali riddles a natural stress-test for LLMs. Carefully constructed, high-signal examples can reveal failure modes invisible to large benchmarks (Levesque et al., 2012; McCoy et al., 2019; Naik et al., 2018), motivating our probe-set design.

## 3 BengaliFig

We describe our methodology for BengaliFig dataset construction in this sections. Our steps are described below:

### 3.1 Data Collection and Preprocessing

We built the BengaliFig corpus by scraping riddles from blogs, forums, and digital archives, then filtering and cleaning them through a compact three–stage pipeline: *deduplication*, *normalization*, and a final *manual audit*.

#### 3.1.1 Deduplication

To ensure that every item is unique yet representative, we combined automatic retrieval with human checks. Each riddle question is a Unicode string $q_i$. For every pair $(q_i, q_j)$ we compute the normalized Levenshtein distance

$$d(q_i, q_j) = \frac{\text{lev}(q_i, q_j)}{\max(|q_i|, |q_j|)} \in [0, 1], \quad (1)$$

where $\text{lev}(\cdot, \cdot)$ is the minimal edit count. Pairs with $d(q_i, q_j) \leq \tau$ were flagged as candidates, starting with $\tau = 0.10$ for high precision and gradually relaxed to 0.30 for recall. Flagged pair was automatically deduplicated only if the answers were perfect overlaps. Within each candidate cluster, we kept the element with smallest identifier as the canonical form. Native speakers then reviewed all the remaining candidates to discard duplicates. This hybrid design delivered near–perfect precision while capturing subtle paraphrases. Our initial Collection consisted of 770 entries. 238 were removed after the automatic and manual deduplication.

#### 3.1.2 Answer Normalization

We standardized answer text by removing extraneous punctuation and isolating the core answer when sources contained extra explanation. For a raw answer $\alpha$ and delimiter set $\mathcal{S} = \{\text{":", "–", "|", "—"}\}$ we define, $\hat{\alpha} = \text{first\_split}(\alpha, \mathcal{S})$, logging them $\mathcal{E} = \{(\alpha, \hat{\alpha})\}$ for manual audits.

#### 3.1.3 Manual Audit

Finally, two native speaker auditors performed a full pass to catch residual issues with answer normalization edits, mistranslations, malformed riddles, hidden duplicates, or question–answer mismatches. Unsalvageable entries were removed; ambiguous but valuable riddles were lightly edited to maintain fluency and logical consistency. Total of 97 entries were removed after manual audit.

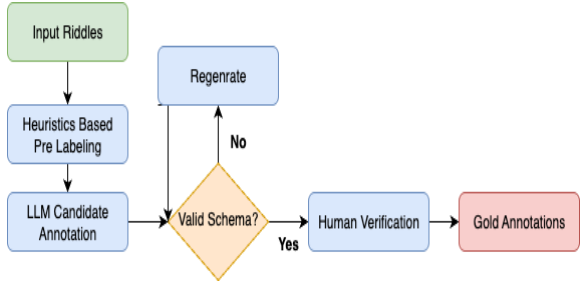

Figure 1: LLM-assisted annotation pipeline. Heuristic priors $p_0$ seed LLM predictions $\hat{y}$, validated against label set $\mathcal{Y}$ and finalized as $y^*$ by human annotators.

## 3.2 LLM-Assisted Human Annotation

Annotating riddles is challenging because solutions hinge on culture, figurative language, and multi-step reasoning. Each riddle is labeled along five orthogonal dimensions, making purely manual work costly. We therefore adopt an **LLM-assisted framework** in which a large language model proposes candidate labels that are then verified and, if necessary, corrected by human annotators. We describe our annotation schema below where we just list down the five orthographic dimensions and their set of possible values. In Appendix A.7 we provide the annotation schema with detailed explanation of the labels alongside the LLM prompt.

**Annotation Schema.** The five dimensions capture complementary cognitive and cultural properties:

**Reasoning Type**
($r \in \mathcal{R}$): { *metaphorical*, *commonsense*, *descriptive*, *wordplay*, *logical_deduction*, *compound* }

**Trap Type**
($t \in \mathcal{T}$): *surface_literal*, *multiple_valid*, *culturally_specific*, *linguistic_trick*, *misdirection*, *archaic_reference*, *none*.

**Cultural Depth**
($c \in \mathcal{C}$) : {*universal*, *cultural_specific*}.

**Answer Type**
($a \in \mathcal{A}$): *place*, *person*, *animal*, *plant*, *object*, *natural_phenomenon*, *body_part*, *food_drink*, *concept*, *quantity*, *text_symbol*.

**Difficulty**
($d \in \mathcal{D}$) : {*easy*, *medium*, *hard*}.

All sets $\mathcal{R}, \mathcal{C}, \mathcal{T}, \mathcal{A}, \mathcal{D}$ are mutually exclusive and exhaustive.

**Framework:** The annotation pipeline, illustrated in Figure 1, proceeds in three stages described below.

**Stage 1: Heuristic Based Pre-Labeling.** Given a riddle–answer pair $(q, \alpha)$, we first compute a vector of heuristic priors $\mathbf{p}_0 \in [0, 1]^{|\mathcal{A}|}$ for the *answer_type* label using regex patterns and gazetteer look-ups derived from Bengali morphology. For example, if $\alpha$ contains suffixes like "পুর/নগর" (pur/nagar) or matches any token in the lexicon set form places, $\mathcal{L}_{\text{place}}$, we set $\mathbf{p}_0[\text{place}] = 1$. Similar detectors exist for animals, plants, body parts, natural phenomena, etc. These lightweight priors injected into the prompt to stabilize the LLM generation.

**Stage 2: LLM Candidate Annotation.** Let the complete label space be $\mathcal{Y} = \mathcal{R} \times \mathcal{A} \times \mathcal{D} \times \mathcal{T} \times \mathcal{C}$, covering reasoning type, answer type, difficulty, trap type, and cultural depth. The LLM (DeepSeek V3) receives $(q, \alpha, \mathbf{p}_0)$ and a compact schema prompt, and must output a candidate annotation for each $(q, \alpha)$ pair as a single valid tuple in strict JSON format: $\hat{y} = (r, a, d, t, c) \in \mathcal{Y}$. Temperature is fixed at $\tau = 0.1$ to minimize randomness. A validator enforces type constraints; any invalid $\hat{y}$ triggers re-prompting with the same $\mathbf{p}_0$. Cost effective inference API was the deciding factor in choosing the DeepSeek for suggesting annotation. LLM's task is not to provide final annotation but suggestions in structured JSON schema which is is easy to for annotators to edit and provide the final annotation, saving time.

**Stage 3: Human Verification** Two native-speaker annotators receive a set of entries to annotate. We first test inter annotator agreement on a 5% (22 out of 435 items) stratified set, and obtained Krippendorff's alpha = 0.9034. (Comprehensive calculation in Appendix A.1). The obtained score is well above the acceptable threshold (0.85) to continue. The remaining 413 riddles were single-annotated after establishing sufficient agreement. The annotators inspect each candidate annotation $\hat{y}$ and either accept it or supply a corrected gold label $y^*$, producing the final gold-standard labels as illustrated in Figure 2. This hybrid design substantially reduces annotation effort while retaining reliability. We also observed that annotation time reduced from $\approx 7.3$ minutes (manual) to $2.4$ minutes per riddle. This human-in-the-loop design preserves cultural fidelity while reducing average annotation time.

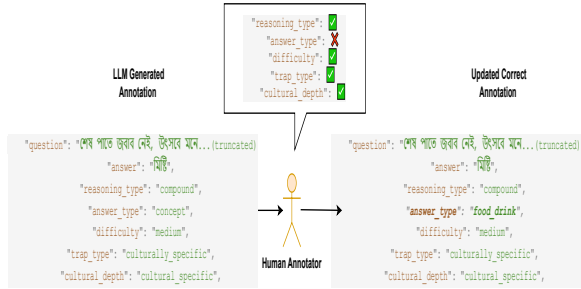

Figure 2: Example of Human annotator providing final validation to LLM generated candidate annotations

## 3.3 Exploratory Data Analysis

Figure 4 summarizes the distribution of the five annotation dimensions across all 435 riddles. **Reasoning type** is dominated by *metaphorical* riddles (224, 52%), which further underscores the contribution of our dataset providing a test bench for Bengali figurative and metaphorical reasoning. **Answer types** are diverse: tangible *objects* leading (127) while culturally salient categories are also well represented. Difficulty skews toward the middle, with *medium* items forming the majority (235), only 8 questions rated *hard*, and the rest *easy*. The riddle style is largely *surface–literal* in its **trap type** (358), with smaller pockets of *linguistic_trick* (57). Majority riddles require **cultural knowledge**, with *cultural_specific* depth accounting for 285 instances.

In Figure 3 *surface–literal* traps occur across both cultural depths but are strongly concentrated in *cultural_specific* items (213 vs. 145), whereas *linguistic_trick* riddles are almost exclusively cultural (56 of 57), highlighting that deceptive wordplay is closely tied to Bengali linguistic nuance. We also observe that riddles requiring wordplay (75 of 83) and compound (63 of 72) reasoning tend to be culturally specific. While commonsense reasoning (31 of 42) is more universal.

Cross–label analyses (Figure 5) reveals that reasoning complexity correlates with difficulty: over half of *compound* riddles are *medium* and a notable 7 are the only cluster of *hard* questions. Metaphorical riddles tend to be more inclined towards medium difficulty, whereas *commonsense* riddles remain predominantly *easy*. The figure on the right illustrates that culturally specific riddle tends to be more difficult than the ones that can be solved with universal basic knowledge.

## 3.4 MCQ Format Creation

Drawing inspiration from (Zellers et al., 2019) we leverage LLMs to generate distractors and to create multiple-choice questions (MCQs). Using a fully automated, two-stage AI pipeline designed to balance *diversity* in candidate distractors with *precision* in final selection. The pipeline consists of 3 steps:

**Step 1: Constraint extraction and prompt conditioning.** Many Bengali riddles state explicit surface clues such as required grapheme count (অক্ষর *akṣar* 'grapheme') or properties like size, color, number, shape, or time. We apply rule-based detectors that (a) identify Bengali numeral words and Bengali digits (১, ২, ৩) when answer's required grapheme count is mentioned, and (b) flag other additional cues such as size, color, shape, count, time etc. These constraints are packed into a structured prompt and attached to each riddle together with its question, answer, reasoning type ($r \in \mathcal{R}$), and answer type ($a \in \mathcal{A}$).

**Step 2: Constraint and misdirection-aware generation.** We give the prompt with extracted constraints to a generator LLM. The generator produces $n = 6$ distractors that exploit the riddle's surface misdirection rather than copying the correct answer. Candidates must: (i) seem plausible under the surface meaning, (ii) sound natural to Bengali speakers, and (iii) follow all constraints like grapheme length. We use higher sampling temperature for diverse outputs.

**Step 3: Automated selection under explicit criteria.** A separate selector LLM ranks candidates using five criteria: misdirection power, first-instinct appeal, surface-logic coherence, constraint compliance, and diversity of traps. The selector uses lower temperature for stable results. We apply basic checks, shuffle options, and record the correct answer's position.

**Model heterogeneity:** We use two different models in our pipeline for practical and methodological reasons. DeepSeek-V3 serves as the generator, while GPT-4 handles the selection task. This division separates the generation and evaluation processes to reduce self-endorsement bias. The choice of DeepSeek for generation was driven by cost considerations and its accessible API. For the selection stage, we chose GPT-4 due to its established reliability in evaluation tasks and consistent

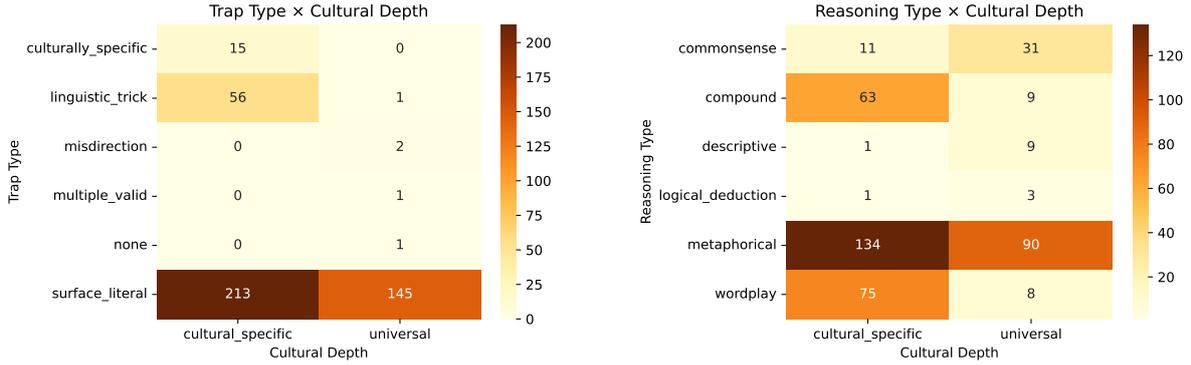

Figure 3: Cross–label relationships: Trap type vs. cultural depth(left) and Reasoning vs. cultural depth(right)

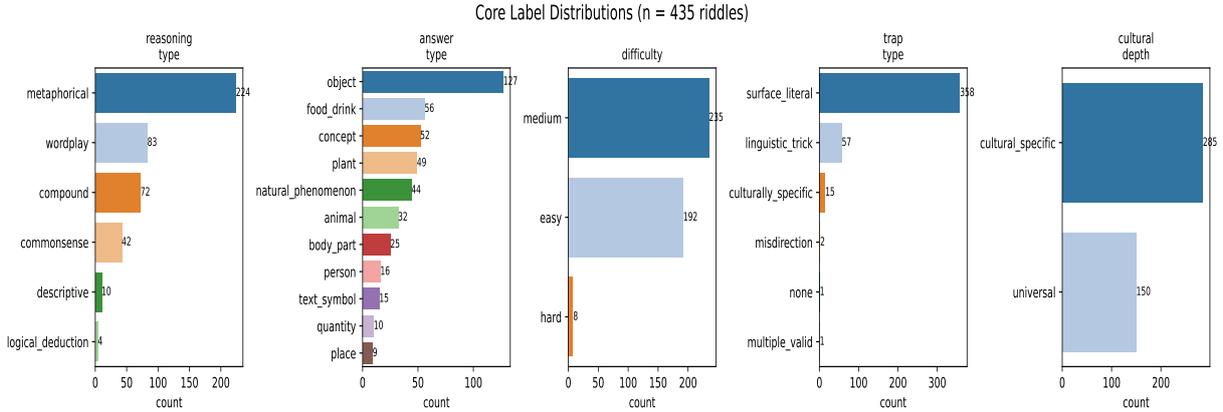

Figure 4: Core label distributions across the five annotation dimensions.

performance across different tasks. (Full prompts in Appendix A.2)

**Observations** In Bengali graphemes represent syllables. Even with constraint blocks, generated distractors often ignored grapheme count requirement. Not a single riddle with a grapheme constraint received a complete set of options meeting that limit. The failure of LLMs to generate distractor options conforming to the grapheme count reveals their graphemic and phonological weakness in non Latin scripts. In contrast, answer-type constraints (e.g., country, fruit, language) were largely respected.

## 4 Experiments and Results

To rigorously assess the figurative reasoning capabilities of large language models (LLMs) on BengaliFig, we developed a comprehensive and robust evaluation framework. This framework is designed to handle multiple model providers, support diverse evaluation modes (zero-shot, and few shot chain-of-thought prompting (CoT)), and guarantee reproducibility through systematic result logging and metadata tracking.

We evaluate a diverse set of LLMs across major providers: (i) **OpenAI**: GPT-4.1 and GPT-5, (ii)**Anthropic**: Claude Sonnet 4.0 and Claude Opus 4.1, (iii)**DeepSeek**: DeepSeek-V3.1[2], (iv)**Meta**: LLaMA-4 Maverick, LLaMA-4 Scout, (v)**Qwen**: Qwen3-235B

### 4.1 Zero-Shot Evaluation

We first assess all models in a strict zero-shot setting, where each riddle is presented with four multiple-choice options and models must return only the single correct letter (A-D). Accuracy is reported over the entire 435-item test set and key annotation dimensions. See Appendix A.3 for prompt and result reproducibility.

**Overall Performance Rankings.** Table 1 presents the comprehensive performance hierarchy. **GPT-5 achieves the highest accuracy at 82.3%**, followed closely by **Claude-Opus-4.1 at 79.8%**, establishing a clear top tier. Performance then drops substantially to GPT-4.1 (69.0%), creating a notable 10.8-point gap that suggests qualitative differences in reasoning capabilities. The remaining models cluster in the 55-63% range, with Claude-Sonnet-4.0 performing weakest at 50.8% and barely exceeds random chance in our 4-option multiple-choice format.

---
[2]DeepSeek-V3.1 was used for final evaluation, upgrading from V3 used in earlier steps.

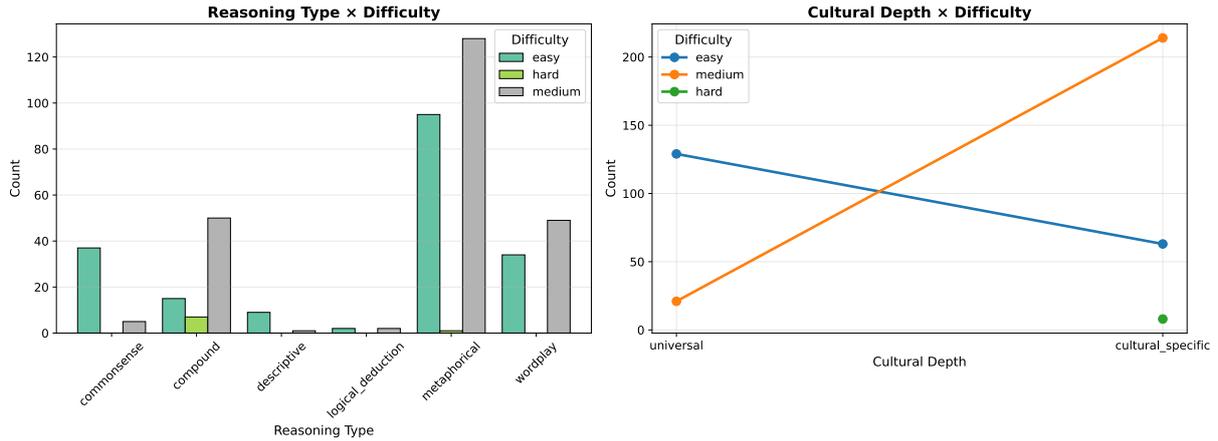

Figure 5: Cross–label relationships: reasoning type vs. difficulty (left); and cultural depth vs. difficulty (right)

| Model | Overall Acc. |
|---|---|
| GPT-5 | **82.3** |
| Claude-Opus-4.1 | 79.8 |
| GPT-4.1 | 69.0 |
| LLaMA-4 Maverick | 63.2 |
| DeepSeek-V3.1 | 59.8 |
| Qwen3-235B | 58.6 |
| LLaMA-4 Scout | 55.2 |
| Claude-Sonnet-4.0 | 50.8 |

Table 1: Zero-shot overall accuracy (%)

**Reasoning Type Breakdown.** Figure 6 reveals pronounced variation in accuracy across reasoning categories. All models perform well on *descriptive* and *logical deduction* tasks, with top performers achieving perfect accuracy (100%) while, *metaphorical reasoning* poses was more challenging. Even for leading models such as GPT-5 and Claude-Opus plateau around 80-81%, suggesting inherent difficulty in abstract conceptual mapping within the Bengali cultural context. *Wordplay* emerges as the most discriminative category, where performance gaps exceed 40 percentage points. GPT-5 leads at 84.3%, while Claude-Sonnet-4.0 achieves only 39.8%. This significant gap underscores the linguistic sophistication required for Bengali phonetic and orthographic manipulation, where models must simultaneously process sound patterns, semantic ambiguity, and cultural references. *Commonsense* and *compound* reasoning showed intermediate difficulty levels, with top models reaching 81-83% accuracy.

**Trap Type.** Surface-literal traps dominate the dataset (358 of 435), so any apparent correlation between trap susceptibility and overall accuracy may be confounded by the class imbalance; detailed analysis is provided in Appendix A.6. Our analysis of trap-type correlations is only exploratory.

**Difficulty and Cultural Depth Analysis.** Tables 2 and 3 reveal systematic performance patterns across BengaliFig's annotation dimensions. Difficulty levels show clear stratification: accuracy decreases monotonically from Easy (Mean: 70.1%, Range: 47.4–85.9%) to Medium (61.7%, 53.2–80.0%) to Hard (29.7%, 0.0–62.5%). The substantial 40.4-point mean gap between Easy and Hard categories validates our annotation scheme while demonstrating genuine cognitive challenges. Even GPT-5 achieves only 62.5% on Hard riddles. Cultural depth analysis reveals consistent but more subtle effects: universal riddles outperform cultural-specific counterparts across all models, with a mean advantage of 10.0 percentage points. This systematic disparity (ranging from +5.6 for GPT-5 to +21.7 for GPT-4.1) indicates that cultural knowledge requirements impose additional cognitive load beyond linguistic competence alone. Notably, the cultural gap persists even for extensively multilingual models, suggesting deeper pragmatic understanding challenges rather than surface-level cultural fact retrieval limitations.

**Grapheme–Constraint Evaluation.** Some riddles explicitly specify that the correct answer must contain a fixed number of Bengali graphemes, a cue that humans can easily exploit to eliminate implausible options. Despite explicitly prompting this constraint, none of the LLMs produced a full set of distractors adhering to the grapheme counts, effectively turning these riddles into elimination tasks for humans. Across this 28–item subset, GPT-5 achieved the highest accuracy (85.7%), followed by GPT-4.1 (60.7%) and Claude-Opus-4.1 (42.9%). All other models, including Claude-Sonnet-4.0 (28.6%), Qwen3 (32.1%), DeepSeek-V3.1 (35.7%), and LLaMA-4 variants (7.1–42.9%) performed much worse. This sharp degradation suggests that current LLMs, further reinforces that even strong multilingual ones, struggle to interpret or consistently respect orthographic and phonological constraints in non-Latin scripts.

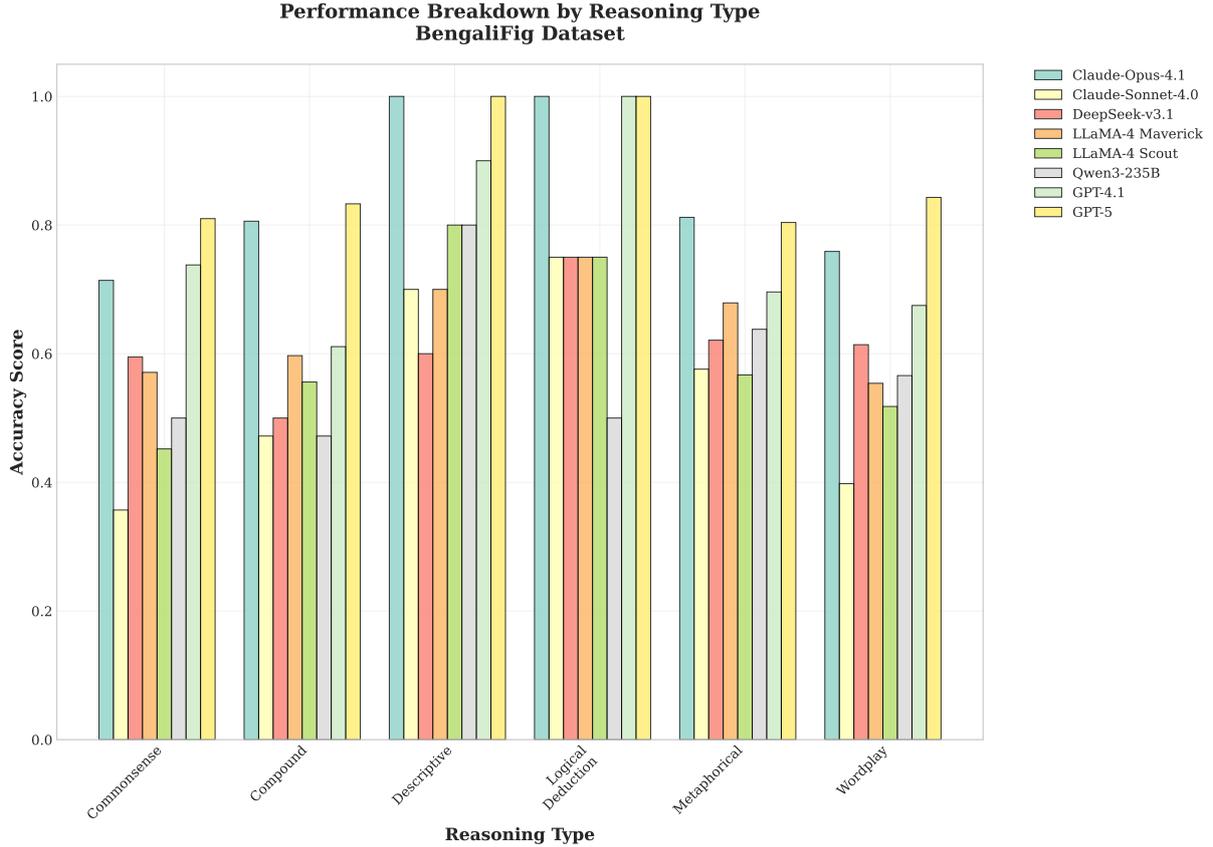

Figure 6: Performance breakdown by reasoning type, revealing significant variation across cognitive categories.

**Proprietary API Cost.** Considering API usage cost in proprietary models, despite leading in accuracy, GPT-5 and Claude-4.1-Opus incur several-fold higher API costs than GPT-4.1(Appendix A.4).

## 4.2 Few-Shot Chain-of-Thought Analysis

To investigate whether structured reasoning can improve performance on challenging riddles, we conduct few-shot Chain-of-Thought (CoT) evaluation on a strategically selected subset of BengaliFig. We identify the subset of "hardest yet solvable" instances which are riddles where exactly one model succeeded in zero-shot evaluation while all others failed. This ensured that our analysis focuses on genuinely difficult but not impossible reasoning challenges. See Appendix A.5 for prompt and reproducibility notes.

| Model | Easy | Med | Hard |
|---|---|---|---|
| GPT-5 | 85.9 | 80.0 | 62.5 |
| Claude-Opus-4.1 | 84.4 | 77.0 | 50.0 |
| GPT-4.1 | 78.6 | 62.6 | 25.0 |
| LLaMA-4 Maverick | 69.3 | 58.7 | 50.0 |
| DeepSeek-V3.1 | 69.8 | 53.2 | 12.5 |
| Qwen3-235B | 66.7 | 54.0 | 0.0 |
| LLaMA-4 Scout | 58.9 | 54.0 | 0.0 |
| Claude-Sonnet-4.0 | 47.4 | 54.0 | 37.5 |
| **Mean** | **70.1** | **61.7** | **29.7** |
| **Std Dev** | 13.7 | 10.1 | 24.1 |

Table 2: Zero-shot accuracy (%) by difficulty level on BengaliFig.

**Experimental Design.** Our few-shot CoT prompt provides two Bengali riddle exemplars with explicit reasoning traces, followed by a structured three-step methodology: (1) riddle type identification and question analysis, (2) systematic option evaluation, and (3) logical conclusion formation. This framework encourages models to decompose complex reasoning while maintaining cultural and linguistic authenticity through native Bengali instruction.

| Model | Cultural | Universal |
|---|---|---|
| GPT-5 | 80.4 | 86.0 |
| Claude-Opus-4.1 | 78.6 | 82.0 |
| GPT-4.1 | 63.5 | 79.3 |
| LLaMA-4 Maverick | 57.5 | 74.0 |
| DeepSeek-V3.1 | 54.0 | 70.7 |
| Qwen3-235B | 54.4 | 66.7 |
| LLaMA-4 Scout | 52.3 | 60.7 |
| Claude-Sonnet-4.0 | 50.2 | 52.0 |
| **Mean** | **61.4** | **71.4** |
| **Std Dev** | 11.4 | 11.1 |

Table 3: Zero-shot accuracy (%) by cultural depth on BengaliFig.

| Model | Zero-Shot | CoT | Gain(%) |
|---|---|---|---|
| GPT-4.1 | 0.0 | 30.0 | **+30.0** |
| Claude-Opus-4.1 | 20.0 | 43.3 | +23.3 |
| DeepSeek-V3.1 | 3.3 | 26.7 | +23.3 |
| LLaMA-4 Maverick | 6.7 | 26.7 | +20.0 |
| LLaMA-4 Scout | 3.3 | 20.0 | +16.7 |
| Qwen3-235B | 6.7 | 23.3 | +16.7 |
| Claude-Sonnet-4.0 | 16.7 | 26.7 | +10.0 |
| GPT-5 | 43.3 | 43.3 | 0.0 |

Table 4: Few-shot Chain-of-Thought performance (%) on hardest yet solvable BengaliFig subset (n=30).

**Differential CoT Efficacy.** Table 4 reveals striking heterogeneity in CoT responsiveness across model families. **GPT-4.1 demonstrates the most substantial improvement**, achieving a 30% accuracy gain (0% → 30%) with 9 successful corrections out of 30 initially failed cases. **Claude-Opus-4.1 and DeepSeek-Chat both achieve 23.3% improvement rates**, though from different baselines, Claude-Opus from a stronger initial position (20% → 43.3%) and DeepSeek from near-zero performance (3.3% → 26.7%).

Conversely, **GPT-5 shows zero improvement with CoT** with 0% despite starting from the highest baseline (43.3%). This counterintuitive finding suggests that GPT-5's zero-shot reasoning may already be near-optimal for this task difficulty level, with CoT providing redundant rather than complementary processing.

**Baseline Performance and CoT Ceiling Effects.** An inverse relationship emerges between zero-shot accuracy and CoT improvement. Models with low initial performance gain most, while stronger models show diminishing returns. This ceiling effect may indicate that CoT mainly helps bridge basic reasoning gaps rather than refine high-level reasoning. We caution that this finding is based on a limited 30-item subset. Across models, accuracy improvements remain below 30%, with no system exceeding 43.3%, aligning with the "hardest yet solvable" design and underscoring the cognitive difficulty of culturally grounded Bengali riddles.

**Final Insights.** Few-shot CoT results reveal that (1) structured reasoning aids mid-tier models but yields limited benefit for top-tier ones, and (2) cultural–linguistic reasoning challenges persist despite explicit reasoning cues. These patterns suggest that deeper cultural grounding, not additional prompting, is key to advancing performance on BengaliFig.

## 5 Conclusion

We introduced **BengaliFig**, a small but carefully constructed challenge set for probing figurative and culturally grounded reasoning in Bengali. Our 435 riddles are annotated along five orthogonal dimensions and converted to multiple–choice format through an AI–assisted pipeline. Evaluation of eight frontier LLMs shows that even state-of-the-art systems struggle, especially with metaphorical and culturally specific riddles. Few-shot chain-of-thought prompting yields only limited gains, confirming diminishing returns for explicit reasoning guidance. A focused analysis of riddles containing explicit Bengali grapheme-count clues reveals a further weakness: most models ignore simple phonological constraints in this non-Latin script, leading to sharp accuracy drops. These findings highlight persistent gaps in cross-lingual and script-aware reasoning and underscore the need for resources that emphasize depth and cultural specificity rather than scale. Although BengaliFig is a text-only resource, many riddles evoke inherently multimodal reasoning, linking linguistic metaphor with perceptual and sensory imagery. Future extensions could therefore explore how multimodal models grounded in language, vision, and sound handle such culturally embedded reasoning tasks in low-resource contexts. We release the data, prompts, and scripts to support future work on figurative and culturally informed evaluation in low-resourced languages.

## Limitations

Our current design focuses solely on textual reasoning, although many riddles implicitly reference visual, auditory, or tactile attributes that future multimodal extensions could capture. Our design as a focused challenge set introduces several constraints. First, the probe set is small (435 riddles), which limits statistical power for fine-grained comparisons and cannot cover the full range of Bengali figurative language. Second, although each item is annotated along five dimensions with native-speaker verification, annotation was performed by only two annotators. After a small pilot to check inter-annotator agreement, the remaining data were split between them rather than double-annotated, so agreement estimates are limited and some subtle labels may reflect individual judgment. Third, our evaluation of few-shot chain-of-thought (CoT) prompting was restricted to a curated subset of the hardest but solvable riddles. This provided useful evidence that CoT helps mid-tier models but does not significantly raise overall reasoning ability, yet running few-shot CoT across the entire probe set could yield additional insights. Fourth, we did not obtain a human performance baseline. Although we planned a small user study to compare human solvers with LLMs, participation relied on voluntary sign-ups and we did not receive enough responses to draw meaningful conclusions.

## Ethics Statement

All riddles were collected from publicly available Bengali websites and digital archives. We removed entries containing personally identifiable information or offensive content and included only items suitable for open research release. Two native speakers performed the annotations after a small pilot to check inter-annotator agreement.

The dataset is released solely as an evaluation resource. Its small size makes it unsuitable for training large models, but it could still be misused to overstate cultural competence. We therefore document its scope and limitations and encourage responsible use in research on figurative reasoning and cross-lingual evaluation.

# A  Appendix

## Contents

### A.1 Inter–Annotator Reliability Calculation

To quantify inter–annotator agreement on the 5 % stratified audit set, we computed Krippendorff's $\alpha$ across all five annotation dimensions jointly.

**Data.** The audit set contained 22 riddles, each annotated along 5 independent dimensions, yielding $N = 22 \times 5 = 110$ annotation units. Each unit was labeled by two annotators.

**Krippendorff's $\alpha$.** For nominal data,

$$\alpha = 1 - \frac{D_o}{D_e},$$

where $D_o$ is the observed disagreement and $D_e$ is the expected disagreement under chance.

Because the five dimensions differ in category counts ($K_1 = 6$, $K_2 = 7$, $K_3 = 2$, $K_4 = 11$, $K_5 = 3$), the expected disagreement is the mean of the per–dimension maxima:

$$D_e = \frac{1}{5} \sum_{i=1}^{5} \left(1 - \frac{1}{K_i}\right) \approx 0.7532.$$

**Observed disagreement.** Across the $N = 110$ units the annotators disagreed on $d = 8$ units, so the observed disagreement is

$$D_o = \frac{d}{N} = \frac{8}{110} \approx 0.0727.$$

**Reliability.** Substituting into the formula,

$$\alpha = 1 - \frac{D_o}{D_e} = 1 - \frac{0.0727}{0.7532} \approx \mathbf{0.9034}.$$

**Interpretation.** Following Krippendorff's guidelines ($\alpha \geq 0.80$ for reliable conclusions), the obtained $\alpha = 0.9034$ indicates **high agreement**. Therefore it is scientifically acceptable to proceed with the planned *non–overlapping* annotation of the remaining dataset.

**Per–dimension statistics.** Table A1 reports per–dimension disagreement counts and Krippendorff's $\alpha_i$ values on the 5% (22 items) stratified audit set. The per–dimension disagreements ($d_i$) sum to the eight total disagreements reported in the main text. Expected disagreement $D_{e,i}$ for each dimension was computed under the maximal–disagreement assumption for nominal categories, $D_{e,i} = 1 - \frac{1}{K_i}$, and per–dimension reliabilities were obtained as $\alpha_i = 1 - D_{o,i}/D_{e,i}$. The joint reliability across all five dimensions is $\alpha \approx 0.90345$, consistent with the value reported in the main text and indicating high inter–annotator agreement.

### A.2 Prompt Templates Used for MCQ Format Creation

#### A.2.1 Distractor Suggestion Prompt

> You are an expert in Bengali riddles and psychological misdirection. Your task is to create {n} clever distractors that exploit the riddle's intended misdirection.
>
> RIDDLE: {question}
>
> CORRECT ANSWER: {answer}
>
> STRATEGY: Bengali riddles work by misdirecting the reader toward an obvious but wrong interpretation. Your distractors should capitalize on this misdirection, NOT be similar to the correct answer. Focus on the main question the riddle seems to be asking at first glance.
>
> ANALYSIS FRAMEWORK:
>
> 1. SURFACE INTERPRETATION: What does the riddle seem to be asking about at first glance?
>
> 2. MISDIRECTION TRAP: What category of answers would most people naturally think of?
>
> 3. COGNITIVE BIAS: What assumptions does the riddle want people to make?
>
> DISTRACTOR CREATION RULES:

Table A1: Per-dimension agreement breakdown on the 5% (22 items) audit set.

| Dimension | $K$ | Units | Disagreements $d_i$ | $D_{o,i} = d_i/22$ | $D_{e,i} = 1 - 1/K$ | $\alpha_i = 1 - D_{o,i}/D_{e,i}$ |
|---|---|---|---|---|---|---|
| Reasoning Type (D1) | 6 | 22 | 3 | 0.1364 | 0.8333 | 0.8364 |
| Trap Type (D2) | 7 | 22 | 2 | 0.0909 | 0.8571 | 0.8939 |
| Cultural Depth (D3) | 2 | 22 | 0 | 0.0000 | 0.5000 | 1.0000 |
| Answer Type (D4) | 11 | 22 | 1 | 0.0455 | 0.9091 | 0.9500 |
| Difficulty (D5) | 3 | 22 | 2 | 0.0909 | 0.6667 | 0.8636 |
| **All** | – | **110** | **8** | **0.07273** | **0.75325** | **0.90345** |

1. Focus on the first main question the riddle seems to ask at first glance.

2. Create answers that fit the OBVIOUS interpretation.

3. Make them plausible for someone who hasn't realized the trick.

4. Include answers from the category people would FIRST think of.

5. Add answers that sound logical but miss the linguistic trick.

6. Avoid answers similar to the correct answer—they must be from different domains.

7. Make someone think "that makes sense" before they realize the trick.

8. The answers must be strictly in বাংলা with no other scripts or languages.

CONSTRAINT REQUIREMENTS (included only when detected by the code):

- CRITICAL: All distractors MUST have exactly {constraints.syllable_count} syllables in Bengali.

- The correct answer "{answer}" has {constraints.correct_syllables} syllables.

- Count carefully: নদী = 2, সাগর = 3, পাহাড় = 3, বাংলাদেশ = 5

- Additional constraints may appear: size, color, shape, time references, etc.

EXAMPLE THINKING PROCESS:

- If the riddle appears to ask about countries, generate country names.

- If it appears to ask about animals, use animal names.

- If it appears to ask about objects, use object names.

REQUIREMENTS:

1. Output distractors strictly in বাংলা with no explanations.

2. Focus on misdirection rather than similarity.

3. Ensure cultural appropriateness for Bengali speakers.

4. Follow detected syllable/letter constraints.

5. Create cognitive traps, not semantic matches.

Output format:

DISTRACTOR_1: বাংলা শব্দ
DISTRACTOR_2: বাংলা শব্দ
…
DISTRACTOR_{n}: বাংলা শব্দ

### A.2.2 Distractor Selection Prompt

You are an expert in cognitive psychology and Bengali riddles. Select the 3 MOST DECEPTIVE distractors that trap people in the riddle's misdirection.

RIDDLE: {question}

CORRECT ANSWER: {answer}

SUGGESTED DISTRACTORS: 1. বিকল্প
2. বিকল্প
3. বিকল্প
…

SELECTION STRATEGY:

Choose distractors that create the strongest cognitive traps, NOT the ones most similar to the correct answer.

EVALUATION CRITERIA:

1. **Misdirection Power**: How well does it exploit the riddle's surface interpretation?

2. **First Instinct Appeal**: Would this be a typical initial guess?

3. **Cognitive Trap Strength**: How convincing is it before someone realizes the trick?

4. **Surface Logic**: Does it make immediate sense?

5. **Diversity**: Prefer distractors from different trap categories.

AVOID:

- Distractors that are too similar to one another.

- Distractors close to the correct answer.

- Obscure or implausible options.

PRIORITIZE:

- Obvious category-based guesses.

- Immediately logical answers.

- Options that delay the "aha!" moment.

Output format:

SELECTED: [comma-separated numbers of the most deceptive options]

### A.3 Zero–Shot Evaluation Prompt and Reproduction Guide

#### A.3.1 Full Prompt Template

All models were queried in Bengali with a single-turn user message. For each riddle the script replaces `{question}` and `{options}` with the actual text and candidates (A–D). The prompt is shown below exactly as sent to the API.

PROMPT:

নিচের ধাঁধাটি সমাধান করুন এবং সঠিক উত্তরের এক অক্ষরে (A, B, C, অথবা D) দিন:

প্রশ্ন: question

বিকল্পসমূহ: A) option_1 B) option_2 C) option_3 D) option_4

শুধু JSON আকারে উত্তর দিন. কোনো ব্যাখ্যা বা বর্ণনা দেবেন না. উদাহরণস্বরূপ: {"উত্তর": "<আপনার উত্তর এখানে>"}

**English Translation**

Solve the following riddle and give the correct answer as a single letter (A, B, C, or D):

Question: question

Options: A) option_1 B) option_2 C) option_3 D) option_4

Provide the answer only in JSON format. Do not include any explanation or description. For example: {"Answer": "<your answer here>"}

The script enforces a temperature of 0 (except where a provider disallows it) and does not include a system message so that every model receives the same pure zero-shot query.

#### A.3.2 Reproducibility Notes

To reproduce the reported zero-shot results:

- **Environment.** Python 3.10+ with the `openai` client library and a valid API key for each provider. Store keys as environment variables (`OPENAI_API_KEY`, `ANTHROPIC_API_KEY`, `NOVITA_API_KEY` etc.).

- **Dataset.** Use the released MCQ JSON file , where each entry contains the riddle, four options, the correct option letter, and the five annotation dimensions.

- **Execution.** Run the provided script and set the provider/model names in the `settings` list. The script automatically handles batching, rate limits, and result logging.

- **Outputs.** For every model a timestamped JSON file is created under `results/zero_shot/`, containing raw model responses, extracted predictions, and per–dimension accuracy statistics.

### A.4 Proprietary Model Evaluation Cost on Zero Shot

Table A2 represents a breakdown of usage cost of proprietary models. Although GPT-5 and Claude-4.1-Opus lead the pack in performance, they come with a significant cost which is several magnitude higher than GPT-4.1 which placed third in overall accuracy. With GPT-5 especially expensive because its lengthy chain-of-thought outputs generate many reasoning tokens that count toward usage fees.

| Model | Cost |
|---|---|
| GPT-4.1 | 0.13$ |
| GPT-5 | 3.40$ |
| Claude-4-Sonnet | 0.43$ |
| Claude-4.1-Opus | 2.19$ |

Table A2: API usage cost of proprietary models

We use a cloud service provider to run evaluation on the open models to reduce infrastructure overhead. However, that is totally optional, as open models are available for free to download and use. As a result, they are not part of API cost analysis.

### A.5 Few–Shot Chain-of-Thought Prompt and Reproduction Guide

#### A.5.1 Full Prompt Template

For the hardest but solvable subset of riddles we used a Bengali few–shot chain-of-thought (CoT) prompt that first presents worked examples and then requests a step-by-step analysis before giving the final answer. Below is the exact template; the three Bengali examples remain fixed, while {question} and {options} are replaced at run time.

PROMPT:

"""

আপনি একটি বাংলা ধাঁধার বিশেষজ্ঞ। নিম্ন কিছু ধাঁধার উদাহরণ দেওয়া হলো, যেখানে ধাঁধার সমাধান বিশ্লেষণ করা হয়েছে।

উদাহরণ ১: প্রশ্ন: একটা ঘড়ির উপর দিয়ে একটা ঘোড়া চলে গেল, ঘড়িটার কটা বাজবে। বিকল্পসমূহ: A. সাতটা B. বারোটা C. নটা D. তিনটা যুক্তি: ঘড়ির কাটা ভেঙে যাবে, তাই বারোটা বাজবে। উত্তর: B

উদাহরণ ২: প্রশ্ন: কোন কার চলে না? বিকল্পসমূহ: A. নৌকা B. সাইকেল C. কুকার D. গাড়ি যুক্তি: কুকার যানবাহন নয়, তাই কুকার চলতে পারে না। উত্তর: C

উদাহরণ ৩: প্রশ্ন: নাকের ডগায় পৈতে আটকান চৈতনে মার টান গলায় ধরে দাও পটকান... বিকল্পসমূহ: A. হামানদিস্তা B. লাটু C. হাতুড়ি D. দা যুক্তি: বর্ণনা লাটুর বৈশিষ্ট্যের সাথে মিলে। উত্তর: B

এখন নিচের ধাঁধাটি সমাধান করুন। প্রথমে যুক্তি ব্যাখ্যা করুন, তারপর উত্তর দিন:

প্রশ্ন: question

বিকল্পসমূহ: options

নিম্নলিখিত ধাপগুলো অনুসরণ করুন: ১. প্রশ্ন বিশ্লেষণ ২. প্রতিটি বিকল্প মূল্যায়ন ৩. যুক্তিসঙ্গত সিদ্ধান্ত

JSON আকারে উত্তর দিন: {"যুক্তি": "<আপনার যুক্তি এখানে>", "উত্তর": "<A/B/C/D>"}

"""

Each model received this full text as a single user message, preceded by a system instruction:

আপনি একটি বাংলা ধাঁধা বিশেষজ্ঞ। সর্বদা ধাপে ধাপে চিন্তা করুন এবং JSON ফরম্যাটে উত্তর দিন।

Temperature was set to 0 when supported.

**English Translated Prompt:**

You are an expert in Bengali riddles. Below are some example riddles with analyses of their solutions.

Example 1: Question: A horse passes over a clock—what time will the clock show? Options: A. Seven o'clock B. Twelve o'clock C. Nine o'clock D. Three o'clock Reasoning: The clock's hands will break, so it will show twelve o'clock.[3] Answer: B

Example 2: Question: Which "car" does not move? Options: A. Boat B. Bicycle C. Cooker D. Car Reasoning: A cooker is not a vehicle, so it cannot move. Answer: C

Example 3: Question: "Tie the thread to the tip of the nose, pull it with force, and let it spin around the neck …" Options: A. Mortar and pestle B. Spinning top C. Hammer D. Machete Reasoning: The description matches the characteristics of a spinning top. Answer: B

Now solve the following riddle. First explain your reasoning, then provide the answer:

Question: question

Options: options

Follow these steps: 1. Analyze the question 2. Evaluate each option 3. Make a logical conclusion

Give the answer in JSON format: {"Reasoning": "<Your reasoning here>", "Answer": "<A/B/C/D>"}

---

[3]Bengali Idiom knowledge is required to understand. In Bengali when it's 12'0 clock for someone or something that means the person or object is in ruins.

The translated prompt contains few-shot examples with reasoning. However, we must mention that, the translated prompts are given only for transparency. A lot of linguistic and cultural essence of these examples are lost in translation.

### A.5.2 Reproducibility Notes

To reproduce the few-shot CoT results:

- **Environment.** Python 3.10+ with the `openai` client library. Store API keys in environment variables (`OPENAI_API_KEY`, `ANTHROPIC_API_KEY`, `NOVITA_API_KEY` etc.).

- **Dataset.** Use the released the curated JSON subset of riddles identified as "hard but solvable" based on zero-shot accuracy.

- **Execution.** Run the provided script and edit the `settings` list to specify provider, model name, and rate-limit delays. The script automatically saves JSON results with raw reasoning, extracted answers, and accuracy statistics.

- **Outputs.** Each run produces a timestamped file in `results/chain_of_thought_hard_cases` containing the full model reasoning text and the parsed predictions, enabling direct comparison with the zero-shot evaluation.

## A.6 Trap Type Analysis

**Trap Susceptibility and Reasoning Robustness.** Figure A1 reports model susceptibility to surface-literal misdirection. Although such traps dominate the dataset (358 of 435 riddles), the comparison is still informative for understanding how models handle superficial cues. Claude-Sonnet-4.0 shows the highest vulnerability (47.6%), whereas GPT-5 and Claude-Opus remain lower at 18.7% and 18.4% respectively.

**Performance–Trap Relationship.** As illustrated in Figure A2, overall accuracy and surface-literal susceptibility exhibit a strong negative correlation ($r = -0.89$). Because the surface-literal category is heavily over-represented, this association should be viewed as exploratory rather than conclusive. Nevertheless, the trend hints that models achieving higher accuracy also develop more robust semantic representations that help them resist superficial distractors. We include these results to encourage further, controlled analyses of the relationship between trap type and reasoning depth.

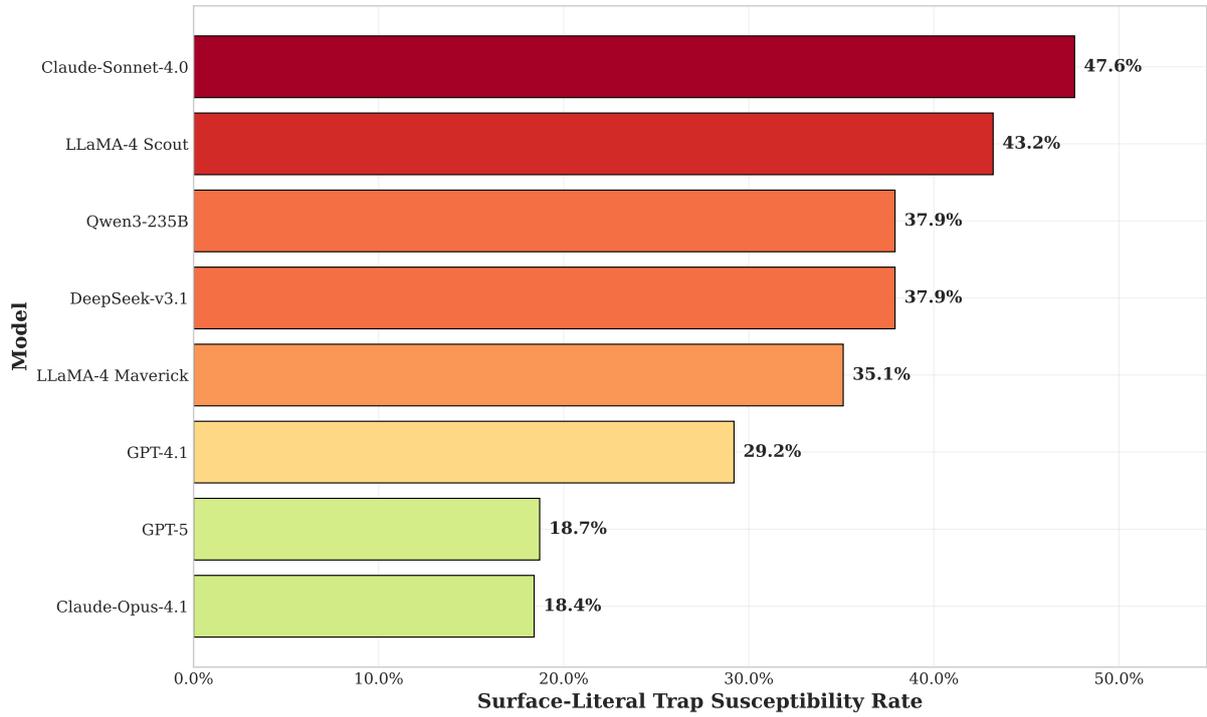

Figure A1: Surface-literal trap susceptibility analysis showing model robustness to misdirection.

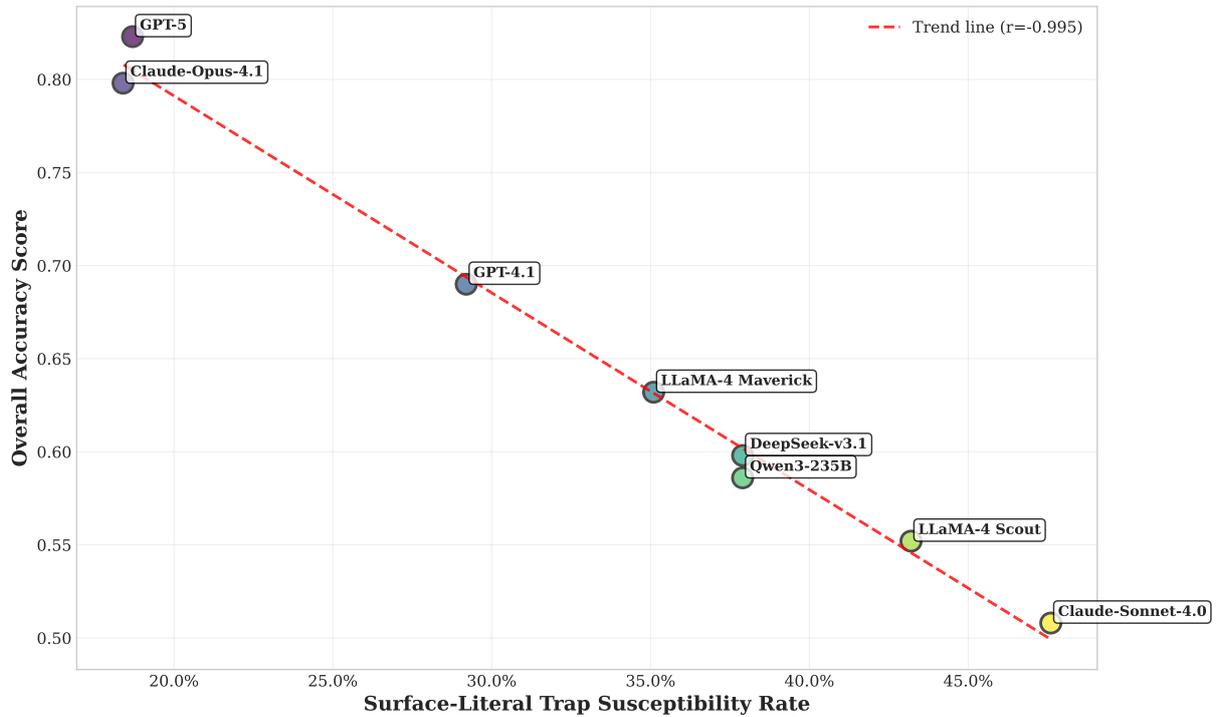

Figure A2: Overall performance versus trap susceptibility, revealing strong negative correlation ($r = -0.89$).

## A.7 LLM Assisted Riddle Annotation

This appendix provides comprehensive details of our LLM-assisted annotation framework used to label the BengaliFig dataset. The annotation process involved two native Bengali speakers who verified and corrected LLM-generated candidate labels across five orthogonal dimensions. We present the complete annotation guidelines provided to human annotators, followed by the exact prompt template used with DeepSeek V3 to generate structured annotation suggestions.

### A.7.1 Annotation Guidelines

The annotation guidilines provided to the annotators are provided in original text followed by English translation below:

**Original Text:** আপনি একটি বাংলা ধাঁধা টীকাকরণ প্রকল্পে অংশগ্রহণ করছেন। আপনার কাজ হলো প্রদত্ত ধাঁধা-উত্তর জোড়াগুলি বিশ্লেষণ করে পাঁচটি নির্দিষ্ট মাত্রায় উপযুক্ত লেবেল প্রদান করা। প্রতিটি ধাঁধার জন্য একটি এআই সিস্টেম প্রাথমিক টীকা সুপারিশ করবে যা আপনি পর্যালোচনা করে প্রয়োজনে সংশোধন করবেন। অনুগ্রহ করে প্রতিটি ধাঁধা সাবধানে পড়ুন, সাংস্কৃতিক প্রসঙ্গ বিবেচনা করুন এবং আপনার সর্বোত্তম বিচারবুদ্ধি প্রয়োগ করুন।

নিম্নলিখিত নির্দেশাবলী অনুসরণ করে বাংলা ধাঁধাগুলির টীকা প্রদান করুন। প্রতিটি ধাঁধা পাঁচটি মাত্রায় লেবেল করতে হবে:

**১. যুক্তির ধরন (Reasoning Type):**

- **রূপক (metaphorical):** প্রতীকী বা রূপক উপস্থাপনা
- **সাধারণ জ্ঞান (commonsense):** দৈনন্দিন জ্ঞান এবং যৌক্তিক অনুমান
- **বর্ণনামূলক (descriptive):** রূপক ছাড়াই আক্ষরিক বৈশিষ্ট্য
- **শব্দখেলা (wordplay):** ধ্বনি প্যাটার্ন, ভাষাগত বৈশিষ্ট্য
- **যৌক্তিক অনুমান (logical_deduction):** সূত্র থেকে ধাপে ধাপে যুক্তি
- **যৌগিক (compound):** একাধিক যুক্তির ধরন একসাথে

**২. উত্তরের ধরন (Answer Type):**

- **স্থান (place):** অবস্থান (আসাম, ঢাকা, বাংলাদেশ)
- **ব্যক্তি (person):** মানুষ, ভূমিকা (মা, শিক্ষক, রাজা)
- **প্রাণী (animal):** উদ্ভিদ ব্যতীত জীবিত প্রাণী (গরু, পাখি, মাছ)
- **উদ্ভিদ (plant):** উদ্ভিদজগৎ (আম, ধান, ফুল)
- **বস্তু (object):** মানবসৃষ্ট জিনিস (কলম, চেয়ার, বই)
- **প্রাকৃতিক ঘটনা (natural_phenomenon):** প্রকৃতির ঘটনা (বৃষ্টি, রোদ, আগুন)
- **শরীরের অংশ (body_part):** শারীরিক অঙ্গ (চোখ, হাত, মুখ)
- **খাদ্য/পানীয় (food_drink):** ভোজ্য পদার্থ (ভাত, পানি, মিষ্টি)
- **ধারণা (concept):** বিমূর্ত ভাব (ভালোবাসা, সময়, স্বপ্ন)
- **সংখ্যা (quantity):** সংখ্যা, পরিমাপ (তিন, শত, মাইল)
- **টেক্সট/প্রতীক (text_symbol):** লিখিত উপাদান (অ, নাম, চিঠি)

**৩. কঠিনতা (Difficulty):**

- **সহজ (easy):** সরাসরি সংযোগ, সাধারণ জ্ঞান
- **মাঝারি (medium):** মাঝারি চিন্তা, কিছু সাংস্কৃতিক জ্ঞান
- **কঠিন (hard):** জটিল রূপক, গভীর সাংস্কৃতিক অন্তর্দৃষ্টি
- **বিশেষজ্ঞ (expert):** অত্যন্ত বিমূর্ত, বিশেষায়িত জ্ঞান প্রয়োজন

**৪. ফাঁদের ধরন (Trap Type):**

- **আক্ষরিক বিভ্রম (surface_literal):** আক্ষরিক ব্যাখ্যা বিভ্রান্তিকর
- **একাধিক বৈধ (multiple_valid):** বেশ কিছু যুক্তিসঙ্গত উত্তর
- **সাংস্কৃতিক নির্দিষ্ট (culturally_specific):** বাঙালি সাংস্কৃতিক জ্ঞান প্রয়োজন
- **ভাষাগত কৌশল (linguistic_trick):** ধ্বনি/শব্দাংশের প্যাটার্ন গুরুত্বপূর্ণ
- **দিক ভ্রষ্টতা (misdirection):** ভুল ইঙ্গিত
- **প্রাচীন উল্লেখ (archaic_reference):** পুরাতন বাংলা শব্দ/উল্লেখ
- **নেই (none):** কোনো উল্লেখযোগ্য ফাঁদ নেই

**৫. সাংস্কৃতিক গভীরতা (Cultural Depth):**

- **সার্বজনীন (universal):** সাধারণ মানবিক জ্ঞানই যথেষ্ট

- **সাংস্কৃতিক নির্দিষ্ট (cultural_specific):** বাংলা ভাষা জ্ঞান বা সাংস্কৃতিক প্রসঙ্গ অপরিহার্য

গুরুত্বপূর্ণ বিষয়:

- বাংলা পাঠ্য সাবধানে পড়ুন, সাংস্কৃতিক প্রসঙ্গ বিবেচনা করুন

- উত্তরের ধরনের জন্য: সবচেয়ে নির্দিষ্ট শ্রেণী ব্যবহার করুন (স্থানের নাম = "স্থান", "বস্তু" নয়)

- সাধারণ বাঙালি বক্তার দৃষ্টিকোণ থেকে কঠিনতা মূল্যায়ন করুন

**English Translation:**

You are participating in a Bengali riddle annotation project. Your task is to analyze the given riddle-answer pairs and provide appropriate labels across five specific dimensions. For each riddle, an AI system will suggest preliminary annotations which you will review and correct as necessary. Please read each riddle carefully, consider the cultural context, and apply your best judgment.

Follow the guidelines below to annotate Bengali riddles. Each riddle must be labeled across five dimensions:

**1. Reasoning Type:**

- **metaphorical:** Symbolic or figurative representation

- **commonsense:** Everyday knowledge and logical inference

- **descriptive:** Literal characteristics without metaphors

- **wordplay:** Sound patterns, linguistic features

- **logical_deduction:** Step-by-step reasoning from clues

- **compound:** Multiple reasoning types combined

**2. Answer Type:**

- **place:** Locations (Assam, Dhaka, Bangladesh)

- **person:** People, roles (mother, teacher, king)

- **animal:** Living creatures except plants (cow, bird, fish)

- **plant:** Vegetation (mango, rice, flower)

- **object:** Man-made items (pen, chair, book)

- **natural_phenomenon:** Nature events (rain, sun, fire)

- **body_part:** Anatomy (eye, hand, mouth)

- **food_drink:** Consumables (rice, water, sweets)

- **concept:** Abstract ideas (love, time, dream)

- **quantity:** Numbers, measurements (three, hundred, mile)

- **text_symbol:** Written elements (letter, name, letter)

**3. Difficulty:**

- **easy:** Straightforward connections, common knowledge

- **medium:** Moderate thinking, some cultural knowledge

- **hard:** Complex metaphors, deeper cultural insight

- **expert:** Highly abstract, specialized knowledge needed

**4. Trap Type:**

- **surface_literal:** Literal interpretation misleads

- **multiple_valid:** Several plausible answers

- **culturally_specific:** Needs Bengali cultural knowledge

- **linguistic_trick:** Sound/syllable patterns matter

- **misdirection:** Red herring clues

- **archaic_reference:** Old Bengali terms/references

- **none:** No significant traps

**5. Cultural Depth:**

- **universal:** General human knowledge sufficient

- **cultural_specific:** Bengali language knowledge or cultural context essential

**Key Points:**

- Read Bengali text carefully, consider cultural context

- For answer type: Use most specific category (place names = "place", not "object")

- Assess difficulty from typical Bengali speaker perspective

**Illustrative Examples** Representative annotations appear in Table A3. For instance, the riddle "কিভাবে কাঁচা ডিম ফেলে কংক্রিটের ফ্লোর ভাঙা যায়, ডিম না ভেঙে?" is labeled (r,t,a,c,d)=(commonsense, surface-literal, concept, universal, easy).

### A.7.2 LLM Annotation Prompt

The following prompt was used with DeepSeek V3 to generate candidate annotations for Bengali riddles. The LLM receives a riddle-answer pair along with heuristic priors and outputs structured JSON annotations that are subsequently verified by human annotators.

PROMPT:
„„„„„„„„„„„„„„„„„„„„„„„„„„„„„„„„„„„„„„„„„„„„„„„„„„„„

```
You are a Bengali riddle expert.
Annotate this Bengali riddle with 5
labels. Focus on the core cognitive
and cultural aspects.
RIDDLE: তিন অক্ষরের এমন দেশ পেট কাটলে খাই যে
বেশ।
ANSWER: আসাম

ANNOTATION SCHEMA:

1. REASONING_TYPE - Primary thinking
required:
```

- metaphorical: Symbolic/figurative representation
- commonsense: Everyday knowledge + logical inference
- descriptive: Literal characteristics without metaphors
- wordplay: Sound patterns, linguistic features
- logical_deduction: Step-by-step reasoning from clues
- compound: Multiple reasoning types combined

```
2. ANSWER_TYPE - What the answer
represents:
```

- place: Locations (আসাম, ঢাকা, বাংলাদেশ)
- person: People, roles (মা, শিক্ষক, রাজা)
- animal: Living creatures except plants (গরু, পাখি, মাছ)
- plant: Vegetation (আম, ধান, ফুল)
- object: Man-made items (কলম, চেয়ার, বই)
- natural_phenomenon: Nature events (বৃষ্টি, রোদ, আগুন)
- body_part: Anatomy (চোখ, হাত, মুখ)
- food_drink: Consumables (ভাত, পানি, মিষ্টি)
- concept: Abstract ideas (ভালোবাসা, সময়, স্বপ্ন)
- quantity: Numbers, measurements (তিন, শত, মাইল)
- text_symbol: Written elements (অ, নাম, চিঠি)

```
3. DIFFICULTY - Cognitive challenge:
```

- easy: Straightforward connections, common knowledge
- medium: Moderate thinking, some cultural knowledge
- hard: Complex metaphors, deeper cultural insight
- expert: Highly abstract, specialized knowledge needed

```
4. TRAP_TYPE - Main misleading element:
```

- surface_literal: Literal interpretation misleads
- multiple_valid: Several plausible answers
- culturally_specific: Needs Bengali cultural knowledge
- linguistic_trick: Sound/syllable patterns matter
- misdirection: Red herring clues
- archaic_reference: Old Bengali terms/references
- none: No significant traps

```
5. CULTURAL_DEPTH - Cultural knowledge
required:
```

- universal: General human knowledge sufficient
- cultural_specific: Bengali language knowledge or cultural context essential

```
SUGGESTED VALUES (verify and adjust):
- answer_type: place

KEY POINTS:
```

- Read Bengali text carefully, consider cultural context
- For answer_type: Use most specific category (place names = "place", not "object")
- Assess difficulty from typical Bengali speaker perspective

```
Output JSON only:
``json
{
"reasoning_type": "...",
"answer_type": "...",
"difficulty": "...",
"trap_type": "...",
```

| Riddle (truncated) | Answer | $r$ | $a$ | $t$ | $c$ |
|---|---|---|---|---|---|
| কাঁচা ডিম ফেলে কংক্রিট... | কোনভাবেই না | commonsense | concept | surface-literal | universal |
| শুইতে গেলে দিতে হয়... | দরজার খিল | compound | object | culturally-specific | cultural-specific |
| সাহেব কোর্ট প্যান্ট পরে... | পেঁয়াজ | metaphorical | food_drink | surface-literal | cultural-specific |
| ১০ জন মানুষ ১০ ঘন্টায়... | কোন সময়ই না | logical_deduction | concept | surface-literal | universal |
| একটা ঘড়ির উপর দিয়ে... | বারোটা | wordplay | quantity | surface-literal | cultural-specific |

Table A3: Sample riddles with final gold annotations $(r, a, t, c)$.

```
"cultural_depth": "...",
"source": "web"
}
` `
```

„„„„„„„„„„„„„„„„„„„„„„„„„„„„„„„„„„„„„„„„„„„„„„„„„„„

We provide here a complete prompt as to how it would be with an example input question-answer pair embedded into it, rather than leaving it as an empty placeholder for transparency and illustrative purpose. The translation and gold annotation after human evaluation is given below:

**Translation Note:** The riddle "তিন অক্ষরের এমন দেশ পেট কাটলে খাই যে বেশ।" translates to "A three-letter country, when you cut its belly, you eat quite well." The answer "আসাম" (Assam) is a wordplay where cutting the middle letter "সা" from "আসাম" gives "আম" (mango), which is eaten. Please note that, Bengali graphemes are compound and represents syllables.

**Gold Standard Annotations:** For the example riddle, the LLM suggested the following annotations which were then verified and modified by human annotators:

- `reasoning_type: wordplay`
- `answer_type: place`
- `difficulty: medium`
- `trap_type: surface_literal`
- `cultural_depth: cultural_specific`